\documentclass[letterpaper]{article}
\usepackage[preprint]{aaai2027}
\usepackage[hyphens]{url}
\usepackage{graphicx}
\usepackage{natbib}
\usepackage{booktabs}
\usepackage{amsmath}
\usepackage{amssymb}
\usepackage{array}

\newcommand{\ours}{Penelope}
\newcommand{\memory}{\mathbf{M}}
\newcommand{\readout}{\mathbf{R}}
\newcommand{\hidden}{\mathbf{H}}
\newcommand{\anchorinit}{\texttt{\char60\kern0.08em\char60}}
\newcolumntype{P}[1]{>{\raggedright\arraybackslash}p{#1}}

\title{Penelope: Localized Latent Recurrence for Efficient Structured Reasoning}
\author{Yutong Chen, Shouqian Shi, Xinran Liu, Haochen Wang,\\
Jiaying Wang, Tianxing Xu, Yuanxi Wang, Zirui Ding}
\affiliations{}

\begin{document}
\maketitle

\begin{abstract}
Complex structured reasoning tasks often require additional computation, yet current language models obtain it mainly by increasing parameter scale or by serializing intermediate steps as chain-of-thought (CoT) tokens. The former raises training and deployment costs, while the latter ties reasoning computation to autoregressive output length. We introduce \ours{}, an efficient latent-reasoning framework for pretrained decoder-only Transformers that localizes recurrent computation to a selected decoder interval. The lower decoder prefix is evaluated once to construct a problem-conditioned boundary memory, which is then iteratively refined through time-modulated GRU dynamics and recurrent readout states before answer generation. A progressive CoT-to-latent curriculum transfers visible reasoning into this internal recurrent path, allowing additional computation to be allocated in latent space without repeatedly executing the complete decoder or generating a long intermediate trace. Experiments on open-source structured-reasoning benchmarks show that, at validation-selected latent budgets, \ours{} attains competitive accuracy relative to established latent-reasoning models while reducing measured inference latency. These results show that latent refinement can be localized to a narrow decoder interval, reducing repeated full-decoder execution without generating a long visible reasoning trace and providing a practical accuracy--efficiency tradeoff for decoder-only Transformer models.
\end{abstract}

\section{Introduction}

Reasoning ability often depends on the amount of computation available for a problem. Current language models obtain additional reasoning computation primarily along two axes. Scaling increases model capacity and inference cost, while also raising training, deployment, and serving costs. Chain-of-thought (CoT) reasoning instead introduces additional autoregressive steps by expressing intermediate computation as generated tokens \citep{wei2023chainofthoughtpromptingelicitsreasoning}. CoT provides an explicit and effective reasoning trace, while coupling the added computation to sequential generation length. Consequently, the two dominant routes bind reasoning computation either to model size or to the number of visible tokens.

Latent reasoning moves intermediate computation from generated tokens into hidden states. Continuous-thought methods show that latent representations can replace portions of a textual trace \citep{hao2025traininglargelanguagemodels,shen2025codi}, while recurrent-depth models expose hidden-state refinement as a test-time compute axis \citep{dehghani2019universaltransformers,geiping2025scalingtesttimecomputelatent}. Many approaches, however, repeatedly execute substantial decoder portions. Their latent trajectories avoid long visible traces, but the marginal refinement cost remains tied to broad model execution. We ask whether comparable reasoning capability can instead be preserved through localized recurrent computation.

Obtaining more reasoning computation and allocating it efficiently are distinct problems. Existing methods largely address the first; we ask how a fixed pretrained backbone can support latent refinement without repeatedly executing the complete decoder. Structured reasoning offers a controlled setting because proof search, graph traversal, and expression reduction naturally evolve compact internal states. Our objective is a computation interface that preserves a problem-specific state while localizing repeated execution at a competitive accuracy--latency operating point.

Realizing such localized latent computation requires addressing three coupled challenges. Sufficient problem-conditioned information must persist without repeated full-decoder execution, motivating a one-time lower-prefix pass and a persistent boundary interface with cached prompt-side context. Reusing only a selected decoder interval must then produce meaningful, stable refinement rather than redundant state updates, motivating fixed-size latent memory, step conditioning, and recurrent updates through a shared output-side interval. Finally, the evolving trajectory must remain accessible to standard answer generation, motivating recurrent readout states and layer-consistent answer-context construction through a final interface pass. These requirements jointly motivate the following design.

We present \ours{}, a localized latent-reasoning framework for decoder-only Transformers. The name draws on Penelope's recurrent weaving: the lower decoder prefix is executed once to establish a stable, problem-conditioned boundary, while a shared output-side interval repeatedly weaves refinements into this persistent latent interface. After a single latent trajectory, the refined interface conditions answer generation. Crucially, \ours{} is not simply a looped decoder block: it computes the prompt prefix once, caches prompt-side context within the selected interval, and thereafter iterates only fixed-size memory and readout states. This organization separates reusable problem contextualization from the computation allocated to refinement, reducing repeated decoder execution without assigning unique reasoning semantics to individual layers.

Our contributions are:
\begin{itemize}
    \item We introduce a cache-compatible latent-reasoning framework that separates one-time prompt contextualization from latent-interface recurrence in a selected decoder interval, avoiding repeated full-decoder execution.
    \item We couple a problem-conditioned boundary memory with recurrent readout states and a layer-consistent answer-context construction, exposing one internal trajectory to standard autoregressive answer generation.
    \item Across three tasks and two decoder-only backbones, matched comparisons maintain competitive reasoning accuracy while reducing sequential decoder-layer applications and measured inference latency.
\end{itemize}

\begin{figure*}[t]
\centering
\includegraphics[width=0.90\textwidth]{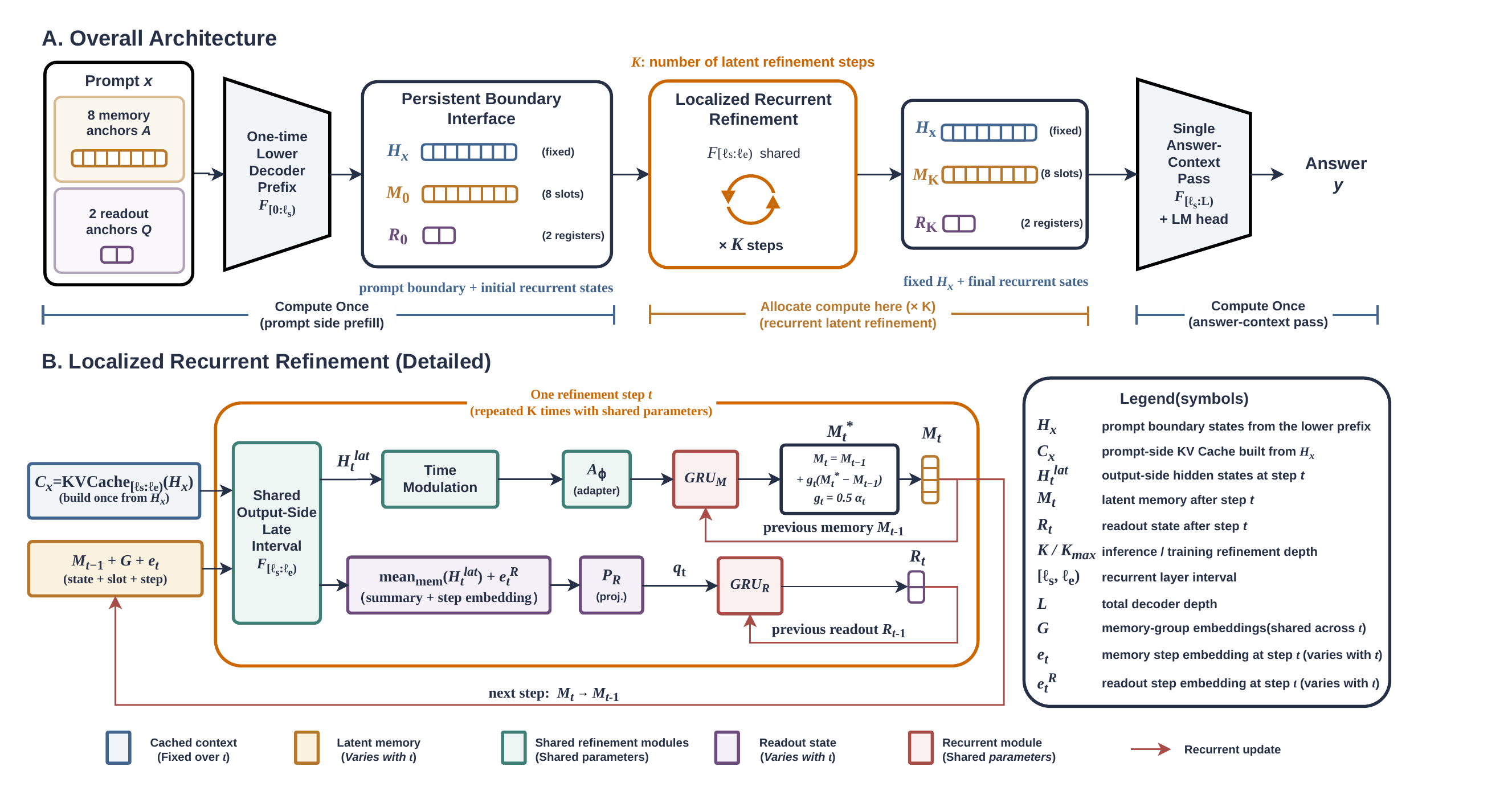}
\caption{Overview of \ours{}. \textbf{A}: a one-time lower prefix forms problem-conditioned boundary states $H_x$, constructs the prompt-side cache $\mathcal{C}_x$, and initializes $(M_0,R_0)$. The shared output-side interval refines only this interface for $K$ steps before one answer-context pass. \textbf{B}: $H_x$ remains fixed across steps, while time-modulated memory and recurrent readout updates produce $(M_t,R_t)$.}
\label{fig:method-overview}
\end{figure*}

\section{Related Work}

\paragraph{Latent reasoning.}
CoT prompting and self-consistency allocate additional inference computation through generated reasoning traces \citep{wei2023chainofthoughtpromptingelicitsreasoning,wang2023selfconsistency}. Coconut progressively replaces textual reasoning steps with continuous states fed back through the decoder, while CODI distills explicit reasoning into implicit trajectories \citep{hao2025traininglargelanguagemodels,shen2025codi}. DART instead distills an autoregressive trace into parallel silent-thought tokens and aligns them through a reasoning-evolvement module \citep{jiang2025dart}. Looped Transformers provide theoretical and empirical evidence that effective recurrent depth can support latent iterative computation without a proportionate increase in distinct parameters \citep{saunshi2025reasoning}. SeLaR takes a complementary selective-computation view, activating soft latent representations at uncertain decoding steps while retaining discrete decoding at confident steps \citep{fu-luo-2026-selar}. Together, these methods establish latent states as a viable reasoning substrate. \ours{} studies the allocation question of how to preserve such an internal trajectory while localizing the decoder computation used to refine it.

\paragraph{Recurrent computation and memory.}
Adaptive Computation Time and PonderNet learn variable internal depth \citep{graves2017adaptivecomputationtimerecurrent,banino2021pondernetlearningponder}, whereas Universal Transformers share a transition across depth \citep{dehghani2019universaltransformers}. Mixture-of-Recursions combines such parameter sharing with lightweight token-level routers that assign dynamic recursive depths \citep{bae2025mixture}. Block-Recurrent and Recurrent Memory Transformers maintain state across blocks or sequence segments \citep{hutchins2022blockrecurrent,bulatov2022recurrentmemorytransformer}. Recent recurrent-depth language models expose latent depth as a test-time compute axis \citep{geiping2025scalingtesttimecomputelatent}. LoopFormer and LoopCoder-v2 further explore looped computation for language and code reasoning \citep{jeddi2026loopformer,yang2026loopcoderv2}. Fixed-Point Reasoners stabilize deep loops with pre-normalization and residual scaling, and use convergence as a learned stopping criterion \citep{movahedi2026fixedpointreasoners}. \ours{} instead adapts a pretrained decoder by coupling a persistent latent interface to recurrent execution within a cached output-side interval; token-level routing and convergence-based halting are complementary to the fixed-depth system evaluated here.

\paragraph{Localized recurrence.}
Existing studies establish that recurrence need not span an entire Transformer. Intra-Layer Recurrence selectively repeats individual layers \citep{nguyen2025intralayerrecurrence}, while Training-Free Looped Transformers damp and reapply a contiguous middle block \citep{chen2026trainingfreeloopedtransformers}. LOTUS instead loops latent blocks and supervises their positions with gold CoT-step tokens \citep{fan2026bridginggaplatentexplicit}. \ours{} advances this direction through a distinct cache-compatible latent computation interface: prompt contextualization is performed once, while only fixed memory and readout states recur within a selected output-side interval before answer generation. The placement analysis evaluates this interval as one efficient operating point; exact placement remains backbone-dependent.

\section{Method}

We consider a decoder-only Transformer with $L$ decoder layers and denote layer $j$ by $F_j$. \ours{} chooses a recurrent interval $[\ell_s,\ell_e)$ of width $r=\ell_e-\ell_s$, giving the decomposition
\begin{align}
F=F_{[\ell_e:L)}\circ F_{[\ell_s:\ell_e)}\circ F_{[0:\ell_s)}.
\label{eq:decoder-decomposition}
\end{align}
This decomposition also gives the architectural meaning of the name \ours{}: $F_{[0:\ell_s)}$ establishes a stable, problem-conditioned foundation once, repeated applications of $F_{[\ell_s:\ell_e)}$ weave successive updates into its persistent latent interface, and the resulting state conditions the final answer-context pass.

\subsection{Localized Latent Interface}

Let $X=E(x)$ denote the prompt embeddings, $A\in\mathbb{R}^{m\times d}$ a fixed-size set of latent anchors, $G\in\mathbb{R}^{m\times d}$ their learned group embeddings, and $Q\in\mathbb{R}^{q\times d}$ the readout anchors. The lower decoder prefix jointly contextualizes the prompt and anchors:
\begin{align}
 [\hidden_x,\memory_0,\readout_0]
   = F_{[0:\ell_s)}([X;A+G;Q]).
 \label{eq:boundary}
\end{align}
Although $A$ is shared before contextualization, causal attention to the prompt makes $\memory_0$ problem-conditioned. The boundary memory serves as a persistent interface between the contextualized problem representation $\hidden_x$ and localized recurrent refinement. These latent states do not represent additional reasoning tokens; they provide fixed internal positions for storing and refining problem-conditioned states. Increasing recurrent depth applies more transformations to this fixed-size interface without extending the latent sequence.

\subsection{Localized Recurrent Refinement}

Latent recurrence commonly increases reasoning depth by re-executing substantial portions of a model. \ours{} instead keeps the reusable representation formed by $F_{[0:\ell_s)}$ fixed and repeatedly applies only $F_{[\ell_s:\ell_e)}$. The cached prefix and recurrent interval thus separate problem contextualization from the decoder workload allocated to refinement: more reasoning computation corresponds to more localized refinement steps, whose marginal decoder depth is $r$ rather than $L$. Whether the boundary remains sufficiently expressive is evaluated by the placement controls in Section~4, rather than assumed from a fixed interpretation of individual layers.

The integer $K$ denotes the number of recurrent refinement steps. The selected
interval first processes the boundary prompt states once and retains their
layerwise key--value tensors:
\begin{align}
 \mathcal{C}_x
   &=\operatorname{KVCache}_{[\ell_s:\ell_e)}(\hidden_x).
 \label{eq:prompt-cache}
\end{align}
In the evaluated output-side configuration, $\ell_e=L$, so
$\mathcal{C}_x$ covers every decoder layer traversed by the final interface.
At step $t\in\{1,\ldots,K\}$, a learned step embedding $e_t$ and the group
embeddings condition the incoming memory. Only these $m$ memory positions are
queried through the selected interval against $\mathcal{C}_x$:
\begin{align}
 \hidden_t^{\mathrm{lat}}
   &= F_{[\ell_s:\ell_e)}
      (\memory_{t-1}+G+e_t;\mathcal{C}_x)_{\mathrm{mem}}.
 \label{eq:lateblock}
\end{align}
Here the semicolon denotes memory queries attending to cached prompt keys and
values. Across steps, prompt-side keys and values remain fixed; only memory
queries and readout states recur. Equation~\ref{eq:lateblock} returns one
feature vector per memory position. To
distinguish successive applications of the shared interval, a time network
receives the normalized coordinates $[t/K_{\max},1/K_{\max}]$ and produces
feature modulation and a positive update scale:
\begin{align}
 (\gamma_t,\beta_t,s_t) &= g_{\tau}([t/K_{\max},1/K_{\max}]),
 \label{eq:time-network}\\
 \widetilde{\hidden}_t
   &= [1+0.1\tanh(\gamma_t)]\odot\hidden_t^{\mathrm{lat}}+0.1\beta_t,
 \label{eq:feature-modulation}\\
 \widehat{\hidden}_t
   &= \widetilde{\hidden}_t+A_{\phi}(\widetilde{\hidden}_t),
 \label{eq:residual-adaptation}\\
 \alpha_t &= 1+0.5\tanh(s_t),
 \label{eq:update-scale}\\
 \widetilde{\memory}_t
   &= \operatorname{GRU}_{M}(\widehat{\hidden}_t,\memory_{t-1}),
 \label{eq:memory-candidate}\\
 \memory_t
   &= \memory_{t-1}+0.5\,\alpha_t
      (\widetilde{\memory}_t-\memory_{t-1}).
 \label{eq:memoryupdate}
\end{align}
Here $\gamma_t,\beta_t\in\mathbb{R}^{d}$ modulate the interval features, $s_t\in\mathbb{R}$ controls the step magnitude, and $A_{\phi}$ is a residual adapter. Equation~\ref{eq:memory-candidate} defines the candidate memory $\widetilde{\memory}_t$, whereas Equation~\ref{eq:memoryupdate} defines the persistent state $\memory_t$ passed to the next refinement. The bounded coefficient $0.5\alpha_t$ interpolates from the incoming memory toward the candidate, allowing the recurrent transformation to change with latent depth while retaining $\memory_{t-1}$ when the proposed correction is small. Figure~\ref{fig:method-overview} abbreviates $0.5\alpha_t$ as $g_t$; its update glyph composes Equations~\ref{eq:feature-modulation}--\ref{eq:memoryupdate} rather than introducing another module.

\subsection{Recurrent Answer Interface}

Additional latent computation can affect the output only if the answer path can access the evolving trajectory. \ours{} therefore maintains answer-accessible recurrent states alongside the boundary memory. Let $\hidden_t^{\mathrm{lat}}$ denote the selected-interval hidden states at the memory positions, $\bar{\hidden}_t^{\mathrm{lat}}=\operatorname{mean}_{\mathrm{mem}}(\hidden_t^{\mathrm{lat}})$ their mean, and $e_t^R$ a readout step embedding:
\begin{align}
 q_t &= P_R(\bar{\hidden}_t^{\mathrm{lat}}+e_t^R)\in\mathbb{R}^{2\times d},
 \label{eq:readout-query}\\
\readout_t &= \operatorname{GRU}_{R}(q_t,\readout_{t-1}).
\label{eq:readout}
\end{align}
Equation~\ref{eq:readout-query} compresses the current memory-position features into two readout proposals, and Equation~\ref{eq:readout} integrates them with the persistent answer state. Every refinement therefore updates a compact state that remains accessible to the final answer path. After $K$ refinements, the final interface $[\memory_K;\readout_K]$ traverses $F_{[\ell_s:L)}$ once as queries against the cached prompt-side keys and values. The resulting interface keys and values are appended to the prompt cache to form one layer-consistent answer context. Standard autoregressive decoding then generates the answer from this cache without any further recurrent refinement. Thus the complete inference path consists of one lower-prefix contextualization, one prompt-side interval cache construction, $K$ latent-interface refinements, one final interface pass, and ordinary cached answer decoding.

\subsection{Progressive Latent Training}

The curriculum transfers computation from explicit token trajectories into the localized latent trajectory. Training starts from the same visible-CoT checkpoint used to initialize the shared-source Coconut baseline. Let a supervised trace contain reasoning steps $(c_1,\ldots,c_J)$ followed by answer $a$. For replacement depth $u$, the target is
\begin{align}
y^{(u)}=
\begin{cases}
(c_{u+1},\ldots,c_J,a), & u<J,\\
a, & u\geq J.
\end{cases}
\label{eq:replacement_target}
\end{align}
Each stage removes one visible reasoning step while adding one latent refinement. The first half of training traverses $K=u=1,\ldots,6$ in equal update blocks; the second half fixes $K=6$ and consolidates answer-only generation. A stage-global pair $(K_s,u_s)$ is shared by every example in a minibatch, and traces shorter than $u_s$ deterministically yield the answer-only target.

For stage $s$, the objective is target-token cross-entropy:
\begin{align}
\mathcal{L}^{(s)}
  =\operatorname{CE}\!\left(
    p_\theta(\cdot\mid x,\memory_{K_s},\readout_{K_s}),
    y^{(u_s)}
  \right).
\label{eq:objective}
\end{align}
Prompt, memory, and readout positions are masked, and the loss is averaged only over non-padding target tokens. The changing target supplies the acquisition signal, while the final stage optimizes the same latent-to-answer path used at inference.

\subsection{Implementation Details}

The experiments use Llama-3.2-1B \citep{dubey2024llama3herdmodels} with $L=16$, the recurrent interval $[\ell_s,\ell_e)=[11,16)$, eight latent interface states, and two readout states. This output-side interval is the evaluated operating point rather than a universal layer optimum. Shared time modulation, residual adaptation, and recurrent updates implement Equations~\ref{eq:lateblock}--\ref{eq:readout}. These modules operate only on the fixed memory/readout positions; they neither extend the prompt nor process generated answer tokens. The final interface pass is included in the pre-answer accounting below, whereas autoregressive answer decoding is excluded. Exact initialization and module hyperparameters are provided in the supplement.

\subsection{Pre-Answer Execution Accounting}

We count \emph{sequential decoder-layer applications} before answer generation, including cache construction and the final interface pass but excluding autoregressive answer decoding. One unit denotes one causally ordered application of one decoder layer to its active positions. It is a serial decoder-depth proxy, not a FLOP count or hardware-independent complexity measure. \ours{} forms the boundary through $\ell_s$ layers, applies the interval of width $r$ once to construct the reusable prompt-side key--value cache, performs $K$ latent-interface refinements through that interval, and passes the final interface through $F_{[\ell_s:L)}$:
\begin{align}
C_{\mathrm{Penelope}}^{\mathrm{preans}}(K)=L+(K+1)r.
\label{eq:compute}
\end{align}
A full-decoder latent-recurrence path with an initial pass, $K$ latent refinements, and a final answer-context pass uses
\begin{align*}
 C_{\mathrm{full}}^{\mathrm{preans}}(K)=(K+2)L.
\end{align*}
Equation~\ref{eq:compute} makes the marginal cost of one additional refinement $r$ rather than $L$, and
\begin{align}
\lim_{K\rightarrow\infty}
 \frac{C_{\mathrm{Penelope}}^{\mathrm{preans}}(K)}{C_{\mathrm{full}}^{\mathrm{preans}}(K)}
 =\frac{r}{L}.
 \label{eq:asymptotic_ratio}
\end{align}
The inference-only $K=0$ diagnostic in Section~4 executes no recurrent update,
but still constructs the boundary and output-side cache and performs the final
answer-context pass. It therefore uses
$C_{\mathrm{Penelope}}^{\mathrm{preans}}(0)=L+r=21$ applications in the
evaluated topology.
This count summarizes serial decoder depth, not aggregate FLOPs: active sequence
length, attention, memory movement, and kernel behavior remain implementation
dependent. The supplement provides cache-aware layer-position accounting and
aligns both topology proxies with synchronized wall-clock measurements.

\section{Experiments}

\subsection{Experimental Setup}

\paragraph{Datasets and metrics.}
Deep ListOps evaluates compositional expression reduction across controlled nesting depths in a task derived from ListOps \citep{nangia2018listopsdiagnosticdatasetlatent,tay2020longrangearena}. The train and validation sets contain 4,800 and 800 expressions, respectively; the 1,600-example test set spans depths 1--12. Answer digits are balanced, and no identifier or normalized-question hash overlaps across the underlying files. Each reasoning trace follows recursive expression-tree reduction and retains the first depth-indexed transitions, followed by the final answer digit.

ProsQA extends the evaluation to synthetic multi-step logical deduction \citep{hao2025traininglargelanguagemodels}, and PrOntoQA evaluates compositional ontology reasoning \citep{saparov2023languagemodelsgreedyreasoners}. ProsQA contains 17,886/300/500 train/validation/test examples; our PrOntoQA split contains 9,000/200/800 examples. Every split passes the same identifier and normalized-question-hash audit. Answer-generation methods decode greedily until EOS or a 32-token cap; Visible CoT uses a 256-token cap. Task-specific extractors are fixed across methods, and empty or nonmatching extractions are incorrect.

\paragraph{Baselines and comparison protocol.}
We compare with Visible CoT \citep{wei2023chainofthoughtpromptingelicitsreasoning}, Coconut \citep{hao2025traininglargelanguagemodels}, CODI \citep{shen2025codi}, and a full-decoder recurrence reference.
On Deep ListOps and ProsQA, \ours{} and Coconut share Llama-3.2-1B, one visible-CoT source checkpoint, data files, three latent-stage runs, effective batch size four, and 1,500 optimizer updates per run. The Visible-CoT and CODI controls use each task's matched continuation-update count and effective batch size while retaining their native targets and objectives. As a cross-backbone compatibility control, we repeat Deep ListOps with Qwen3.5-0.8B-Base \citep{qwen2026qwen35modelcard}. Training disables automatic mixed precision; model-storage dtypes follow the released configurations. Primary accuracy and synchronized latency results use runtime-verified BF16 inference. The source uses rank-16 LoRA \citep{hu2021loralowrankadaptationlarge} ($\alpha=32$, dropout .05) on all attention and MLP projections. \ours{} and Coconut continue these adapters at learning rate $2\times10^{-5}$ and train method-specific modules at $10^{-4}$; CODI retains its projection-and-distillation protocol. Optimization uses AdamW, weight decay .01, cosine decay, 5\% warm-up, and gradient-norm clipping at 1.0.

This design matches the source checkpoint, dataset, base minibatch size, latent-stage runs, optimizer-update count, and data exposure while preserving each method's native objective. It does not parameter-match the methods or equalize training FLOPs, wall-clock time, or target presentations. The pretrained backbone parameter count remains unchanged. Runtime auditing gives Coconut 11.28M trainable parameters, while \ours{} uses 11.27M LoRA plus 64.08M recurrent-module parameters (75.35M total). The latter instantiate the latent memory, readout, and refinement interface, so method-specific trainable capacities and storage footprints differ. Our efficiency claim concerns inference topology and measured latency, not parameter efficiency. The supplement reports the audit and a separately qualified continuation-time estimate.

On Deep ListOps, Coconut uses $K=8$, and CODI uses six implicit states. Validation independently selects the smallest best \ours{} depth for each run, giving $(5,4,2)$. The continuation methods use the same three latent-stage seeds; full-decoder recurrence is evaluated separately through a matched interval analysis. On ProsQA, Coconut ends at six latent steps, while validation selects \ours{} depths $(1,1,6)$.

PrOntoQA uses 3,000 optimizer updates and effective batch size eight for both methods because its traces contain at most six reasoning steps. Coconut uses six latent steps, and \ours{} selects the smallest validation-optimal depth from $K\in\{1,\ldots,6\}$ before test evaluation, giving $(3,3,1)$. All main comparisons use fixed validation-selected depths rather than online input-adaptive halting. We report mean and sample standard deviation over three latent-stage runs drawn from the same visible-CoT source checkpoint. The readout and curriculum controls retain FP32 within each matched comparison, whereas the placement analysis and cross-method rankings use runtime-verified BF16 inference.

\paragraph{Inference timing.}
Unless noted otherwise, latency measures greedy generation with batch size one
and runtime-verified BF16 on an NVIDIA RTX PRO 6000. Inputs are tokenized and
transferred before timing, and CUDA synchronization brackets every generation.
Latency uses paired-checkpoint measurements for direct comparisons or run-level
measurements for averaged reporting.
The specified Deep ListOps pair and all ProsQA runs use 32 warm-up examples per
method; Qwen and PrOntoQA use 16. Deep ListOps and PrOntoQA report synchronized
per-example distributions for one specified checkpoint pair, whereas ProsQA
reports the mean and sample standard deviation of three paired run-level means.
Table~\ref{tab:primary_main} distinguishes these latency objects from the
three-run EM statistic. Hardware, software, and distributional details appear
in the supplement.

\subsection{Accuracy--Efficiency Comparison}

We evaluate whether \ours{} maintains competitive latent-reasoning accuracy while reducing repeated decoder computation.

\begin{table*}[t]
\centering
\scriptsize
\setlength{\tabcolsep}{3.0pt}
\renewcommand{\arraystretch}{0.93}

{\itshape (a) Deep ListOps}\par\vspace{2pt}
\begin{tabular*}{\textwidth}{@{\extracolsep{\fill}}lP{0.30\textwidth}crrr@{}}
\toprule
Method & Inference path & Runs & EM (\%) $\uparrow$ & Output Tokens $\downarrow$ & Latency (ms) $\downarrow$ \\
\midrule
Visible CoT & autoregressive trace & 3 & $51.27{\pm}0.92$ & 69.00 & $1047.46{\pm}63.74^{\ddagger}$ \\
Coconut & 8 full-decoder latent steps & 3 & \textbf{$52.79{\pm}0.36$} & 4.00 & $188.15{\pm}1.02^{\dagger}$ \\
CODI & 6 implicit thought states & 3 & $51.17{\pm}0.31$ & 3.00 & \underline{$161.19^{\mathsection}$} \\
\ours{} & validation-selected fixed $K$ & 3 & \underline{$52.25{\pm}0.72$} & 4.00 & {\bfseries\boldmath $99.82{\pm}0.65^{\dagger}$} \\
\bottomrule
\end{tabular*}
\vspace{3pt}

{\itshape (b) ProsQA}\par\vspace{2pt}
\begin{tabular*}{\textwidth}{@{\extracolsep{\fill}}lP{0.30\textwidth}crrr@{}}
\toprule
Method & Inference path & Runs & EM (\%) $\uparrow$ & Output Tokens $\downarrow$ & Latency (ms) $\downarrow$ \\
\midrule
Visible CoT & autoregressive trace & 3 & $45.87{\pm}0.46$ & 52.07 & $780.69{\pm}187.00^{\ddagger}$ \\
Coconut & 6 full-decoder latent steps & 3 & \textbf{$79.87{\pm}1.80$} & 8.65 & $252.25{\pm}15.94^{\ddagger}$ \\
CODI & 6 implicit thought states & 3 & \underline{$78.27{\pm}1.10$} & 7.65 & \underline{$231.86{\pm}3.42^{\ddagger}$} \\
\ours{} & validation-selected fixed $K$ & 3 & \underline{$78.27{\pm}0.90$} & 8.65 & {\bfseries\boldmath $168.19{\pm}27.70^{\ddagger}$} \\
\bottomrule
\end{tabular*}
\vspace{3pt}

{\itshape (c) PrOntoQA}\par\vspace{2pt}
\begin{tabular*}{\textwidth}{@{\extracolsep{\fill}}lP{0.30\textwidth}crrr@{}}
\toprule
Method & Inference path & Runs & EM (\%) $\uparrow$ & Output Tokens $\downarrow$ & Latency (ms) $\downarrow$ \\
\midrule
Visible CoT & autoregressive trace & 3 & \textbf{$99.71{\pm}0.19$} & 87.79 & $1295.99{\pm}9.33^{\ddagger}$ \\
Coconut & 6 full-decoder latent steps & 3 & $99.58{\pm}0.14$ & 3.00 & $160.71{\pm}0.63^{\dagger}$ \\
CODI & 6 implicit thought states & 3 & $96.00{\pm}2.39$ & 2.00 & \underline{$145.50{\pm}0.38^{\ddagger}$} \\
\ours{} & validation-selected fixed $K$ & 3 & \underline{$99.67{\pm}0.07$} & 3.00 & {\bfseries\boldmath $92.85{\pm}16.70^{\dagger}$} \\
\bottomrule
\end{tabular*}

\caption{Accuracy--latency comparison on three structured-reasoning tasks with Llama-3.2-1B. Penelope maintains competitive exact-match accuracy while reducing inference latency relative to the evaluated full-decoder latent-reasoning path.
    Within each panel, bold and underlined entries denote the best and
    second-best EM, or the lowest and second-lowest latency, respectively.
    EM reports mean${\pm}$sample standard deviation over three continuation runs.
    Latency reports paired checkpoints for direct comparisons or run-level
    measurements for averages; all use batch-1, runtime-verified BF16 generation.
    $\dagger$ marks one
    specified synchronized checkpoint pair; its ${\pm}$ value is the per-example
    timing standard deviation. $\ddagger$ marks mean${\pm}$sample standard
    deviation across three run-level latency means; $\mathsection$ marks one
    run-level mean. Latency statistics are therefore distinct from the EM
    statistics. Penelope and Coconut train approximately 75.35M and 11.28M
    parameters, respectively; the efficiency claim concerns inference execution
    and measured latency, not parameter efficiency.}
\label{tab:primary_main}
\end{table*}

Table~\ref{tab:primary_main}(a) shows that \ours{} maintains competitive reasoning accuracy relative to full-decoder latent reasoning while reducing repeated decoder execution. Under matched optimizer-update and data-exposure budgets, \ours{} reaches $52.25\pm0.72\%$ EM, within 0.54 point of Coconut, while exceeding Visible CoT and CODI by 0.98 and 1.08 points, respectively. Its four-token answer path avoids the long visible trace and repeatedly executes only the five-layer output-side interval. The placement analysis below isolates this computational choice from the method-specific training objectives.

The same localized path transfers to logical deduction and ontology reasoning. On ProsQA, \ours{} obtains $78.27\pm0.90\%$ EM compared with Coconut's $79.87\pm1.80\%$ and CODI's $78.27\pm1.10\%$; the matched-update Visible-CoT reference reaches $45.87\pm0.46\%$. Across three paired checkpoint runs, \ours{} uses validation-selected depths $(1,1,6)$ and records $168.19\pm27.70$ ms, whereas Coconut uses $K=6$ and records $252.25\pm15.94$ ms. Pooling the paired per-example timings gives a 33.3\% relative reduction (paired-bootstrap 95\% CI: 33.0--33.6\%).

On PrOntoQA, Visible CoT reaches $99.71\pm0.19\%$, while \ours{} and Coconut obtain $99.67\pm0.07\%$ and $99.58\pm0.14\%$; the official-aligned CODI transfer reaches $96.00\pm2.39\%$. The specified first continuation checkpoint pair uses \ours{} at $K=3$ and Coconut at $K=6$, requiring $92.85\pm16.70$ and $160.71\pm0.63$ ms, respectively, or 42.2\% lower mean latency. The dispersions describe the per-example timing distributions for that pair, not the three-run EM replication. The selected \ours{} depths $(3,3,1)$ remain fixed before test evaluation.

The principal advantage over full-decoder latent reasoning is inference efficiency. At the specified Deep ListOps checkpoint pair, \ours{} and Coconut use $K=5$ and $K=8$, corresponding to 46 and 160 sequential decoder-layer applications before answer generation. Their synchronized complete-subset means are $99.82\pm0.65$ and $188.15\pm1.02$ ms, a 46.9\% reduction. Unlike the three-run EM, the dispersions summarize per-example timing for this locked checkpoint pair.

\begin{figure*}[t]
\centering
\includegraphics[width=0.49\textwidth]{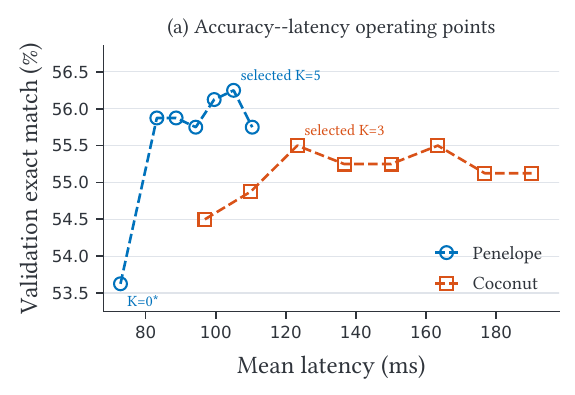}\hfill
\includegraphics[width=0.49\textwidth]{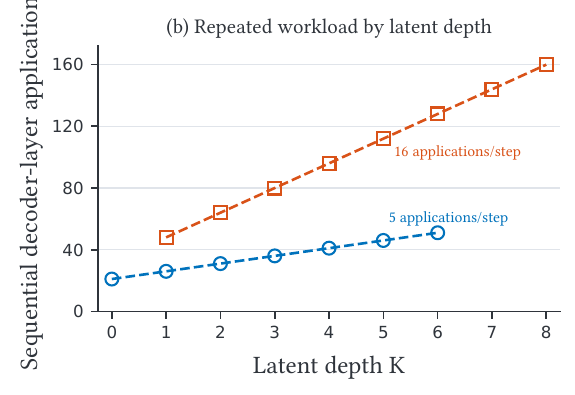}
\caption{Runtime-verified BF16 validation efficiency for one matched Penelope/Coconut checkpoint. (a) Exact match versus synchronized mean latency; annotations mark the smallest validation optima, and $K=0^*$ is an inference-only boundary diagnostic. (b) Sequential decoder-layer applications versus latent depth. Dashed segments connect discrete evaluated budgets.}
\label{fig:accuracy-latency}
\end{figure*}

The Qwen control in Table~\ref{tab:backbone_scale}(a) tests architectural
compatibility under a second decoder-only backbone. In the matched run,
\ours{} reaches 52.06\% EM, within 0.19 point of Coconut and 0.81 point above
CODI. At the paired operating point on the same RTX PRO 6000, \ours{} uses
$K=4$ and requires 207.11 ms, whereas Coconut uses $K=8$ and requires
469.00 ms, a 55.8\% reduction in mean latency. The Visible-CoT and CODI
latencies are separate single-run measurements and do not enter the paired
ratio. All four methods parse 100\% of the 1,600 test examples.

The three-run replication in Table~\ref{tab:backbone_scale}(b) gives
$52.00\pm0.78\%$ EM for \ours{} and $51.38\pm0.78\%$ for Coconut under the
same source and update budget. A zero-shot large-model reference and the
protocol-specific continuation-time estimate are reported separately in the
supplement because neither is a parameter- or timing-matched comparison.

Figure~\ref{fig:accuracy-latency} applies the same smallest-best validation rule to both recurrent paths. Across $K=1,\ldots,6$, Penelope remains within a 0.50-point accuracy band and reaches its maximum of 56.25\% at $K=5$; Coconut first reaches its maximum of 55.50\% at $K=3$, with additional steps increasing cost without improving the selected score. At these operating points, panel (a) reports 105.10 versus 123.28 ms, and panel (b) reports 46 versus 80 sequential decoder-layer applications. Localization therefore reduces the repeated workload by 42.5\% at the validation-selected comparison while maintaining comparable exact-match performance. The curves are discrete budget sweeps rather than evidence of monotonic test-time scaling, and no test prediction participates in depth selection.

\subsection{Mechanism and Placement Analysis}

As a mechanism diagnostic, inference-only $K=0$ isolates the contextualized boundary from
recurrent refinement. Across the three Penelope checkpoints, $K=0$ obtains
$51.56\pm0.81\%$ pooled test EM, whereas independently validation-selected
depths obtain $52.25\pm0.72\%$. The gain is positive in all three runs and
averages 0.69 point; a paired bootstrap over test identifiers gives a 95\%
confidence interval of $[0.27,1.10]$ points. This small, consistent difference
shows that the selected recurrent path can correct some boundary-only
predictions. It is not a training-matched comparison, however: $K=0$ uses the
same recurrence-trained checkpoints rather than a separately optimized
boundary-only model, and the nonmonotonic sweep does not imply that every
additional step improves accuracy.

The transition dynamics are tested as a single architectural unit. Replacing time modulation, the residual adapter, gated integration, and the memory GRU with the direct overwrite $\memory_t=\hidden_t^{\mathrm{lat}}$, while retaining boundary memory and recurrent readout, lowers test EM from $52.25\pm0.72\%$ to $51.44\pm0.35\%$. The paired improvement is positive in all three runs and averages 0.81 point. This control supports regulated state refinement without attributing the gain to an individual MLP, gate, or GRU term; complete per-run values and schedule sensitivity appear in the supplement.

\begin{table}[t]
\centering
\small
\setlength{\tabcolsep}{4.2pt}
\renewcommand{\arraystretch}{0.96}
\begin{tabular*}{\columnwidth}{@{\extracolsep{\fill}}lrrr@{}}
\toprule
Recurrent interval & Layers & Layer calls $\downarrow$ & EM (\%) $\uparrow$ \\
\midrule
Early & $[0,5)$ & 51 & $52.04{\pm}0.47$ \\
Middle & $[6,11)$ & 51 & $52.23{\pm}0.85$ \\
Output-side & $[11,16)$ & 51 & $52.17{\pm}0.97$ \\
Full decoder & $[0,16)$ & 128 & $52.44{\pm}0.81$ \\
\bottomrule
\end{tabular*}
\caption{Matched Deep ListOps placement control at fixed $K=6$ and three
continuation seeds. Only the recurrent interval changes; EM uses
runtime-verified BF16 test evaluation.}
\label{tab:placement-main}
\end{table}

Table~\ref{tab:placement-main} varies only the interval at fixed $K=6$; the
source, interface, transition modules, curriculum, data, update budget, and
continuation seeds remain fixed. Early, middle, and output-side recurrence lie
within 0.19 point, while full-decoder recurrence is only 0.27 point above the
output-side interval despite $2.51\times$ more sequential decoder-layer applications.
The differences are smaller than run-level variation, supporting the
computational sufficiency of localization rather than a universal layer
ranking. Per-run placement results and descriptive layerwise analyses appear
in the supplement.

\begin{table}[t]
\centering
\small
\setlength{\tabcolsep}{3.2pt}
\renewcommand{\arraystretch}{0.96}

{\itshape (a) Matched Qwen3.5-0.8B control}\par\vspace{2pt}
\begin{tabular*}{\columnwidth}{@{\extracolsep{\fill}}lrrr@{}}
\toprule
Method & EM (\%) $\uparrow$ & Output Tokens $\downarrow$ & Latency (ms) $\downarrow$ \\
\midrule
Visible CoT & 50.31 & 125.13 & $3962.08^{\mathsection}$ \\
CODI & 51.25 & 3.00 & \underline{$374.06^{\mathsection}$} \\
Coconut & \textbf{52.25} & 3.00 & $469.00^{\dagger}$ \\
\ours{} & \underline{52.06} & 3.00 & {\bfseries\boldmath $207.11^{\dagger}$} \\
\bottomrule
\end{tabular*}
\vspace{4pt}

{\itshape (b) Three-run replication}\par\vspace{2pt}
\begin{tabular*}{\columnwidth}{@{\extracolsep{\fill}}lcr@{}}
\toprule
Method & Runs & EM (\%) $\uparrow$ \\
\midrule
Coconut & 3 & $51.38{\pm}0.78$ \\
\ours{} & 3 & \textbf{$52.00{\pm}0.78$} \\
\bottomrule
\end{tabular*}

\caption{Qwen3.5-0.8B-Base cross-backbone evidence on Deep ListOps. Panel (a) uses
one matched continuation run and BF16 inference; bold and underlined entries
denote the best and second-best EM, or the lowest and second-lowest latency,
respectively. Its
$\dagger$ entries form a
paired RTX PRO 6000 measurement (\ours{} $K=4$, Coconut $K=8$);
$\mathsection$ marks separate single-run task measurements. Panel (b) reports
mean${\pm}$sample standard deviation across three continuation runs.}
\label{tab:backbone_scale}
\end{table}

\section{Discussion and Future Work}

The evaluation covers three structured-reasoning settings with deterministic answer metrics; detailed depth and placement analyses use compositional expression reduction. This matched-source setting isolates the accuracy--latency tradeoff induced by localization. On Deep ListOps, validation-selected recurrence provides a small, consistent gain over the strong inference-only $K=0$ boundary rather than monotonic improvement with depth. The demonstrated advantage is competitive accuracy with reduced repeated decoder execution. The system selects a fixed validation depth per run; online per-input halting, larger backbones, and more heterogeneous state transitions remain future work.

\section{Conclusion}

We present \ours{}, which allocates latent computation through cache-compatible recurrence over a selected decoder interval. A persistent, problem-conditioned interface separates reusable contextualization from latent-interface refinement, while layer-consistent caches expose the refined trajectory to answer generation. Across three structured-reasoning tasks, \ours{} preserves comparable exact-match performance with less repeated full-decoder execution and lower measured latency. This is efficient latent-computation allocation, not parameter efficiency or monotonic depth scaling.

\clearpage
{\small\bibliography{references}}

\end{document}